\documentclass[sigconf,nonacm]{acmart}
\usepackage{booktabs}
\usepackage{tabularx}
\usepackage{placeins}
\usepackage{tikz}
\usetikzlibrary{arrows.meta,positioning,fit,backgrounds}
\usepackage{balance}

\AtBeginDocument{%
  \providecommand\BibTeX{{%
    \normalfont B\kern-0.5em{\scshape i\kern-0.25em b}\kern-0.8em\TeX}}}

\begin{document}
\raggedbottom

\title{From Location Phrases to Geographic Entities: Task-Adapted Retrieval for People Search}

\author{Yanbo Li}
\author{Chujie Zheng}
\authornote{Work done while at LinkedIn.}
\author{Jiahao Xu}
\author{Chetan Bhole}
\affiliation{%
  \institution{LinkedIn}
  \city{Sunnyvale}
  \state{CA}
  \country{USA}}
\email{yanbli@linkedin.com}

\author{Lingyu Zhang}
\author{\mbox{Puneet Singh Ahluwalia}}
\author{Kevin Nguyen}
\authornotemark[1]
\author{\mbox{Raghavan Muthuregunathan}}
\affiliation{%
  \institution{LinkedIn}
  \city{Sunnyvale}
  \state{CA}
  \country{USA}}
\email{clzhang@linkedin.com}

\author{Santhosh Sachindran}
\author{Sachin Ahuja}
\author{Fedor Borisyuk}
\authornote{Corresponding author.}
\affiliation{%
  \institution{LinkedIn}
  \city{Sunnyvale}
  \state{CA}
  \country{USA}}
\email{ssachindran@linkedin.com}

\renewcommand{\shortauthors}{Yanbo Li et al.}

\begin{abstract}
People search must map free-form location phrases to geographic entities used as
structured retrieval filters. Lexical standardizers handle canonical names well
but are brittle to aliases, misspellings, metropolitan expressions, and
same-name ambiguity. We formulate this task as graded, set-valued entity
retrieval over a fixed ontology. We identify three coupled design requirements:
distinguishing identity-preserving variation from knowledge-dependent aliases,
controlling false negatives among valid same-name entities, and separating
stable transformations from mutable entity knowledge. We realize them in a
prompt-asymmetric bi-encoder with calibrated alias support, bounded
ambiguity-aware negatives, and editable entity documents that support localized
updates without retraining.

Across a fixed production-derived development benchmark and a public GeoNames
transfer task, task adaptation improves substantially over frozen encoders and
standard token baselines. Controlled development ablations show that specialized
supervision contributes beyond standard task fine-tuning and encoder scaling.
On GeoNames, the adapted model improves known-target Recall@1 throughout
zero-to-moderate character overlap, while character $n$-grams retain a small
aggregate Target Recall@5 advantage. In a blinded human comparison on a
stratified production challenge set, our model raises relevant P@1 from 28.0\%
to 46.0\% ($p=0.012$). Fixed-query endpoint estimates improve on non-canonical
queries and remain close to control on frequent queries; a randomized live
experiment detects no engagement regression. These results support
task-adapted geographic entity retrieval as a practical replacement for the
incumbent taxonomy-based standardizer, with the largest relevance gains on
non-canonical queries.
\end{abstract}

\ccsdesc[500]{Information systems~Retrieval models and ranking}
\ccsdesc[300]{Information systems~Users and interactive retrieval}
\ccsdesc[300]{Computing methodologies~Learning latent representations}

\keywords{location grounding, geographic entity retrieval, dense retrieval,
toponym resolution, editable entity representations, people search}

\maketitle

\section{Introduction}
\label{sec:intro}

People search queries combine unstructured intent, such as skills, titles, names,
and companies, with structured constraints. A query such as ``engineers in the
greater Boston area'' requires the location phrase to be mapped to the canonical
entities consumed by candidate generation.
This mapping is commonly implemented as a \emph{geographic standardizer}: given
a location span extracted by query understanding, it returns one or more
identifiers from a curated geographic ontology.

Production standardizers have traditionally combined gazetteers, taxonomy
aliases, typeahead indices, and manually specified string rules. These systems
are interpretable and accurate on high-frequency, well-formed queries, but
coverage declines on less conventional expressions.
Abbreviations (``la'', ``WPB''), informal names (``pink city''), metropolitan
phrasing (``philly metro''), misspellings, and variations in order or punctuation
may not have an exact entry in the alias table. Adding rules improves individual
cases, but requires continuing maintenance and can introduce collisions among
places that share a surface form.

Dense retrieval maps varied surface forms and canonical entities into a shared
space, but this setting raises three coupled design questions. Which query
transformations preserve entity identity, and which aliases require external
validation? How can training expose confusable same-name entities without
penalizing valid alternatives as false negatives? Which transformations belong
in shared model parameters, and which mutable facts should remain in entity
documents? These questions concern positive support, negative sampling, and
memory placement rather than encoder scale alone.

We present a geographic standardizer deployed in the people search stack of a
large professional social network. The system fine-tunes a 0.6B-parameter
instruction embedding model as a shared-weight bi-encoder. Query and entity
inputs use different prompts, while entity representations are precomputed.
For the experiments in this paper, we evaluate a compact representation for
fixed-inventory retrieval.

Our formulation jointly specifies three objects: the support of valid query forms,
the distribution of confusable negatives, and the boundary between stable
transformations in model parameters and mutable knowledge in entity documents.
We formalize these objects in a single fixed-ontology retrieval objective. The design does not require encoder-specific architectural changes and also
applies to fixed-inventory retrieval tasks with non-canonical surface forms,
ambiguous names, and evolving entity metadata.

\noindent\textbf{Contributions.}
\begin{enumerate}
  \item We formulate short-query geographic grounding as graded, set-valued
  retrieval over a fixed, multi-granularity ontology. This formulation captures
  the downstream semantics in which ordered Top-$k$ entities become a
  disjunctive candidate-generation constraint.
  \item We instantiate a unified supervision design with three matched
  components: identity-preserving views expand invariant support; a
  generate--verify--calibrate procedure adds knowledge-dependent aliases without
  overwhelming canonical forms; and dominance-gated, capped same-name negatives
  increase confusability while controlling false negatives.
  \item We separate stable cross-entity transformations in model parameters
  from mutable facts in entity documents, and compare entity-only updates with
  lexical insertion. The comparison exposes complementary update paths:
  entity documents support localized neural updates without retraining, while
  lexical insertion remains stronger for known deterministic mappings.
  \item We evaluate the complete design against standard surface-form fine-tuning
  and frozen encoders up to 8B, test public transfer against lexical baselines on
  GeoNames, and compare directly with the production system in a blinded
  five-annotator study. Online evaluation separates fixed-query relevance from
  live engagement guardrails.
\end{enumerate}

Under a fixed-seed controlled development study, the selected task-adapted
system raises nDCG@10 from 0.8099 under standard surface-form fine-tuning to
0.9201; larger frozen encoders do not close the gap. Public GeoNames transfer provides
independent, judge-free target metrics: the adapted model exceeds exact, prefix,
and BM25 retrieval, improves Target Recall@1 overall and throughout
zero-to-moderate character overlap, and trails character $n$-grams slightly on
aggregate Target Recall@5 because of high-overlap aliases. Entity-document edits
improve held-out-alias retrieval with unrelated controls unchanged, but incur a
canonical-retention cost; direct lexical insertion remains stronger for curated
mappings. On a stratified production challenge set, the model raises
human-judged relevant P@1 from 28.0\% to 46.0\%. Fixed-query endpoint point
estimates rise for non-canonical queries, while the live experiment detects no
engagement regression.

\section{Related Work}
\label{sec:related}

\textbf{People search and structured query understanding.}
Professional people search combines free-text intent with structured constraints
and must retrieve and rank profiles at scale \cite{geyik2018talent}. Recent
systems move beyond lexical matching: Gupta et al. simplify queries and member
documents, fine-tune an embedding model, and use Matryoshka representations for
semantic profile retrieval \cite{gupta2025people}; Borisyuk et al. combine query
understanding, embedding retrieval, an LLM relevance judge, and a distilled
reranker \cite{borisyuk2026semantic}. These systems retrieve profiles from the
user's overall intent. We study a complementary query-understanding stage:
after a location span is extracted, it must be grounded to a fixed geographic
ontology whose ordered Top-$k$ entity IDs become a disjunctive retrieval
constraint. The output is therefore an entity set rather than a member ranking,
and several geographic granularities can be simultaneously relevant.

\textbf{Toponym resolution and geographic representation.}
Toponym resolution conventionally detects place mentions in documents and links
them to gazetteer entries, using context to resolve names such as ``Melbourne''
\cite{gritta2018toponym,weissenbacher2019semeval}. GeoNorm improves candidate
generation and transformer reranking with ontology and population signals
\cite{zhang2023geonorm}; GeoPLACE predicts geographic attributes before
constraining deterministic ontology lookup \cite{zhang2024geoplace}.
Complementary work learns spatial representations from coordinates, map
relations, or nearby entities \cite{li2022spabert,li2023geolm}. Recent retrieval
formulations contrastively encode point-of-interest mentions and gazetteer
entries \cite{nakatani2025poi}, while Masis and O'Connor study noisy,
multilingual, user-provided location strings \cite{masis2024earth}. Our inputs
often contain little context because they consist only of an extracted search span.
The correct output can include several granularities or same-name locations. We therefore
evaluate ordered sets of ontology entities, not only a single coordinate or
gazetteer entry.

\textbf{Entity normalization over alias-rich ontologies.}
Dense entity retrieval provides the closest non-geographic precedent. Dual
encoders retrieve knowledge-base entities without an alias-table candidate
generator \cite{gillick2019entity}, and BLINK scales this formulation to
million-entity linking \cite{wu2020blink}. Biomedical normalization makes the
alias problem explicit: BioSyn learns from incomplete synonym sets through
synonym marginalization \cite{sung2020biosyn}, whereas SapBERT aligns aliases
belonging to the same ontology concept \cite{liu2021sapbert}. Recent work also
systematically studies label verbalization and negative sampling in dual-encoder
disambiguation \cite{rucker2025verbalized}. Our setting combines three
properties: queries are short and frequently misspelled, knowledge-dependent
aliases may be absent from the ontology, and one surface can have several valid
outputs with different relevance grades. This motivates separate supervision
for identity-preserving variation, validated alias acquisition, and bounded
same-name disambiguation instead of treating every unlabeled entity as negative.

\textbf{Synthetic supervision and negative construction.}
LLM-generated queries can turn demonstrations into task-specific retriever
training data; Promptagator additionally filters generated pairs using
round-trip consistency \cite{dai2023promptagator}. Dense retrieval also benefits
from mined negatives \cite{xiong2021ance}, but unlabeled positives make
unfiltered hard-negative mining unreliable. RocketQA addresses this issue with
cross-encoder-denoised negatives \cite{qu2021rocketqa}. Our
generate--verify--calibrate procedure instead targets entity-specific
geographic aliases: generation expands knowledge support, verification checks
the referent, and calibrated sampling prevents synthetic forms from displacing
canonical queries. Likewise, our negatives use ontology identity, hierarchy,
and a bounded prominence relation rather than equating retrieval rank with
irrelevance.

\textbf{Compact retrieval and editable entity memory.}
Dual encoders make candidate representations precomputable
\cite{huang2013dssm,karpukhin2020dpr,reimers2019sbert}, and Matryoshka training
makes vector prefixes useful at multiple dimensions
\cite{kusupati2022matryoshka}. Description-based zero-shot entity linking shows
that independently encoded content can support entities unseen during task
supervision \cite{logeswaran2019zero}; retrieval-augmented models more broadly
separate parametric behavior from external memory \cite{lewis2020rag}.
DynamicER studies emerging mentions and evolving entities through continual
adaptation \cite{kim2024dynamicer}. We investigate a narrower update operation
for a fixed ontology: add a held-out alias to the affected entity document,
re-embed only that entity, and leave model parameters fixed. This makes the
knowledge boundary operational rather than merely architectural.

\textbf{LLM relevance assessment.}
LLM judges reduce assessment cost but require explicit validity evidence
\cite{faggioli2023perspectives}. Calibrated assessors can reproduce searcher
preferences, yet remain sensitive to prompt paraphrases
\cite{thomas2024searcher}; UMBRELA reports strong correlations between system
rankings induced by LLM and human judgments \cite{upadhyay2024umbrela}. Such
run-level agreement does not imply interchangeable query--result labels. Across
eight assessors, Fr\"obe et al. find stronger LLM--LLM than LLM--human agreement
and potential preference for LLM-based rankers \cite{froebe2025assessors}.
LARA and LLM-Rubric use human calibration or explicit multidimensional rubrics
to reduce this gap \cite{takehi2025lara,hashemi2024rubric}. SAGE follows this
domain-calibrated direction for People Search \cite{le2026sage}, but does not
validate our prompted geographic-entity judge. We therefore freeze the judge's
model, prompt, rubric, and input fields across systems and directly measure its
agreement with a blinded five-annotator study.

Taken together, prior work addresses contextual toponym linking, alias-rich
entity normalization, synthetic retriever supervision, editable entity memory,
and LLM assessment largely as separate problems. Our contribution is not a new
base encoder or judge. It is the fixed-ontology retrieval formulation that
couples graded, set-valued relevance with validated alias support, bounded
same-name contrast, and editable entity documents, then tests that formulation
through entity-only alias updates, public transfer, blinded production
comparison, and online serving evidence. This combination distinguishes the
work from both end-to-end profile retrieval and single-referent gazetteer
linking.

\section{Problem Formulation}
\label{sec:problem}

Let $\mathcal{E}$ be a curated set of geographic entities. Each entity $e$ has a
canonical name, alternate names, type (e.g., city, metropolitan area,
administrative division, or country), containment hierarchy, and an aggregate
member-count signal. Given a location span $q$ extracted from a people search
query, the standardizer returns an ordered list
$R_k(q)=(e_1,\ldots,e_k)\in\mathcal{E}^k$ of distinct entities. Candidate generation then uses
the corresponding entity IDs as structured filters. The embedding model receives
only the extracted span $q$, not the full people search query.

A location span can admit multiple relevant entities. ``SF'', for example, may
refer to both a city and a surrounding market area. Conversely, an unqualified
name such as ``Alexandria'' requires the dominant interpretation to precede less
probable same-name entities. We therefore model relevance as graded and evaluate
both ranking quality and downstream result coverage.

In the reported experiments, we retrieve $k=5$ entities. Their identifiers are
passed to candidate generation as a disjunction (an OR filter); the embedding
scores determine their retrieval order. For binary Top-$k$ evaluation, a query
is counted as correct when at least one relevant entity appears in
$R_k(q)$. Graded ranking metrics additionally reward placing the most relevant
interpretation earlier while allowing multiple entities to be valid.

Formally, let $y(q,e)\in\{0,1,2,3,4\}$ be the relevance grade defined by
Table~\ref{tab:rubric}. For a returned list $R_k(q)$, we compute
\begin{equation}
 \operatorname{DCG@}k(q)=\sum_{i=1}^{k}
 \frac{2^{y(q,e_i)}-1}{\log_2(i+1)},
 \label{eq:dcg}
\end{equation}
For each reported comparison, let $G_q$ be the deduplicated union of the
Top-10 entities returned by all compared systems for query $q$. Every pair in
this shared pool is judged once. $\operatorname{IDCG@10}(q)$ is the DCG of the
ten highest grades in $G_q$, while each system's DCG uses its own returned
ordering. Thus all systems share the same denominator. nDCG@10 is defined as
zero when the ideal gain is zero. For threshold $\tau\in\{3,4\}$, Success@5 is the fraction of queries
satisfying $\max_{e\in R_5(q)}y(q,e)\geq\tau$. This query-level success
metric asks whether the OR filter contains at least one usable interpretation,
whereas nDCG distinguishes a dominant referent from
a valid but unlikely same-name entity and rewards placing it earlier. Reporting
both avoids collapsing a multi-entity operating point into single-label
accuracy.

We use a shared encoder $h_\theta$ with prompt-dependent inputs:
\begin{equation}
 \mathbf{u}_q=h_\theta(p_q,q), \qquad
 \mathbf{v}_e=h_\theta(p_e,d(e)),
\end{equation}
where $p_q$ is a retrieval instruction, $p_e$ is an identity prompt, and $d(e)$
is an entity document. Vectors are $\ell_2$ normalized and scored by
$s(q,e)=\mathbf{u}_q^\top\mathbf{v}_e$. Retrieval returns the $k$ highest-scoring
entities, optionally after a type or product filter. For the experiments reported here, we evaluate a compact $D=64$
representation with $k=5$ over a large fixed ontology.

Learning requires specifying both the positive support and the exclusion
boundary. Let $p_+(q\mid e;S_e,A_e,\mathcal{T})$ be the distribution of query
forms for entity $e$, induced by canonical forms $S_e$, validated aliases $A_e$,
and identity-preserving transformations $\mathcal{T}$. Let
$p_-(\mathcal{N}\mid q,e)$ generate a negative set from three sources:
hierarchy-overlap entities, dominance-gated same-name entities, and in-batch
entities. The training problem is
\begin{equation}
 \min_\theta\;\mathbb{E}_{e\sim p_\mathcal{E},\,q\sim p_+,\,
 \mathcal{N}\sim p_-}
 \left[\ell_{\mathrm{ctr}}(q,e,\mathcal{N};\theta,A_e)\right].
 \label{eq:unified_objective}
\end{equation}
The method in Section~\ref{sec:method} specifies these two distributions and
the entity document $d(e,A_e)$. Surface invariance and alias verification shape
$p_+$; hierarchy overlap and bounded dominance shape $p_-$; the mutable field
$A_e$ determines which facts can be changed without updating $\theta$. This
separates the proposed learning formulation from the underlying contrastive
bi-encoder architecture.

Two constraints shape the design. First, the lexical incumbent has its highest
accuracy on the high-volume head, whereas the proposed method targets a smaller
tail; evaluation must report these regimes separately. Second, entity documents
change slowly and can be encoded offline. Retrieval therefore requires one
query encoding followed by exhaustive scoring over compact entity vectors,
rather than online encoding of both sides or cross-encoder inference.

\section{Retrieval Architecture}
\label{sec:system}

\begin{figure}[t]
\centering
\begin{tikzpicture}[
  node distance=3mm,
  box/.style={draw, rounded corners, align=center, font=\scriptsize,
              inner sep=3pt, minimum height=6mm},
  offbox/.style={box, fill=black!4},
  onbox/.style={box, fill=black!8},
  arr/.style={-{Latex[length=1.6mm]}, thick},
]
\node[offbox] (corpus) {Canonical geo corpus\\+ validated aliases};
\node[offbox, below=of corpus] (encE) {Shared encoder $h_\theta$\\identity prompt, offline};
\node[offbox, below=of encE] (vecs) {Normalized compact\\entity vectors};
\node[offbox, below=of vecs] (index) {Exact-search index\\precomputed vectors};
\draw[arr] (corpus) -- (encE);
\draw[arr] (encE) -- (vecs);
\draw[arr] (vecs) -- (index);

\node[onbox, right=17mm of corpus] (q) {Extracted location span $q$};
\node[onbox, below=of q] (encQ) {Shared encoder $h_\theta$\\query prompt};
\node[onbox, below=of encQ] (u) {Normalized compact\\query vector};
\node[onbox, below=6.5mm of u] (knn) {Filtered exact Top-$k$};
\node[onbox, below=of knn] (out) {Geographic entity IDs};
\draw[arr] (q) -- (encQ);
\draw[arr] (encQ) -- (u);
\draw[arr] (u) -- (knn);
\draw[arr] (knn) -- (out);
\draw[arr] (index.east) -- ++(0.55,0) |- (knn.west);

\begin{scope}[on background layer]
  \node[draw, dashed, rounded corners, fit=(corpus)(index),
        label={[font=\scriptsize\itshape]above:offline refresh}] {};
  \node[draw, dashed, rounded corners, fit=(q)(out),
        label={[font=\scriptsize\itshape]above:online retrieval}] {};
\end{scope}
\end{tikzpicture}
\caption{Prompt-asymmetric, shared-weight bi-encoder. Entity representations are
precomputed; each query requires one encoding and an exhaustive filtered search.}
\label{fig:arch}
\Description{A shared bi-encoder precomputes geographic entity vectors offline and encodes each location query online before exact filtered retrieval.}
\end{figure}
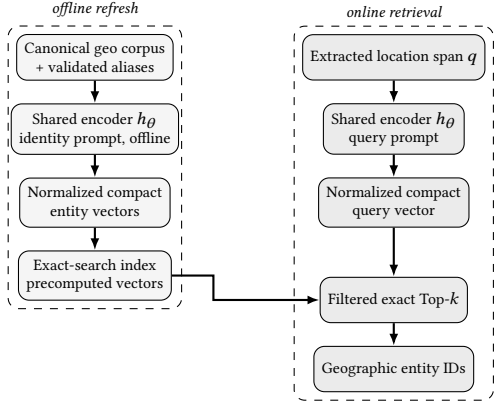

Figure~\ref{fig:arch} separates the weekly entity pipeline from the per-query
path. Offline, we assemble one text document per geographic entity, encode it,
truncate and normalize the resulting vector, and store the vectors in a compact
exact-search index. A compact, low-precision representation permits exact scoring over the fixed
entity ontology.

Online, upstream query understanding supplies only the extracted location span;
the full natural-language people search query is not passed to the embedding
encoder. The serving path encodes the short string into the same compact serving
space, scans the filtered corpus exactly, and returns the Top-$k$ canonical
entity IDs, not free text. At $k=5$, all returned IDs are applied as OR filters by
downstream candidate retrieval; downstream people ranking otherwise remains
unchanged.

Two properties enable exhaustive retrieval. First, the fixed entity side is
precomputed during the weekly refresh, leaving only the query representation to
be computed per request. Second, the training loss directly optimizes the compact serving subspace.
This representation supports exact scoring without approximation-induced recall
loss or an index-specific tuning variable.

Aliases are included in the entity document as well as in query-side training
examples. Section~\ref{sec:alias_updates} defines the resulting entity-only
update, and Section~\ref{sec:update_benchmark} evaluates it: a targeted alias
addition requires re-embedding only the affected entity.

\textbf{Compatibility with the production contract.}
The embedding path replaces only geographic standardization. It consumes the
same extracted span and returns the same geographic entity identifier type as the
incumbent taxonomy-based system, so candidate generation and people ranking do
not require model-specific features. Retrieval depth is an explicit treatment:
Top-1 passes one entity filter, while Top-5 passes five retrieved IDs as a
disjunction. This stable boundary supports three comparisons without changing the rest of
the stack: entity-level offline evaluation, paired calls to the production
endpoint, and a randomized live experiment. It also limits the scope of an
alias edit, because refreshing an entity vector changes geographic matching but
not the downstream ranking model.

\section{Learning Robust and Editable Geo Representations}
\label{sec:method}

\subsection{Design principles}

A generic bi-encoder objective collapses three distinct error channels in
geographic grounding. \emph{Surface variation} changes form without changing
place identity; \emph{knowledge-dependent aliases} such as ``pink city''
cannot be derived from string transformations; and \emph{ambiguity} arises
when one string legitimately denotes several entities. Treating them as
undifferentiated augmentation either misses knowledge-bearing expressions or
creates false negatives for ambiguous names.

We assign each channel to a different part of
Equation~\ref{eq:unified_objective}. Identity-preserving views and validated
aliases define positive support under $p_+$; calibrated sampling allocates that
mass. Hierarchy-overlap and bounded same-name construction shape $p_-$. A
mutable entity-side alias set provides a separate update path: re-embedding an
edited entity changes entity-specific knowledge without updating the encoder.
This allocation assigns each failure mode to a corresponding supervision or
memory mechanism. It does not require encoder-specific architectural changes
and avoids treating the three channels as interchangeable augmentation.

For entity $e$, let $H_e$ denote its canonical name, type, and containment
hierarchy; let $A_e$ be a validated, mutable alias set. We render an entity
document
\begin{equation}
 d(e,A_e)=\operatorname{render}(H_e,b(e),A_e),
 \label{eq:entity_doc}
\end{equation}
where $b(e)=\operatorname{Bucket}(m_e)$ and $m_e$ is an aggregate
prominence signal for $e$. The resulting text contains the
canonical name; city, administrative divisions, and country when available;
entity type; a token representing the popularity bucket $b(e)$; and an
\texttt{Also Known As} field. For
example:
\begin{quote}\small
\texttt{Name: Mountain View, Administrative Division 1: California, Country:
United States, Popularity bucket: b, Type: City, Also Known As: mv ca.}
\end{quote}
The bucket is a coarse platform-specific prominence proxy rather than a census
population estimate. We use natural language for inspectability and
compatibility with the encoder's pretraining format; a later JSON training
variant did not improve development-set retrieval metrics. The final entity
document excludes geohashes: the query contains no corresponding spatial token,
and removing the document-side fields did not materially change development-set
point estimates.

\subsection{Identity-preserving query views}
\label{sec:positives}

For each entity document, we construct a set $S_e$ of query forms conditioned
on entity type. A populated place may yield its name alone, name plus country,
or name plus first-level
administrative division; thus ``San Jose'', ``San Jose US'', and ``San Jose
California'' are distinct anchors for the same entity.

We then sample transformations $t\in\mathcal{T}$ known to preserve the entity
identity and construct $q=t(s)$ for $s\in S_e$. The transformation set includes
country and US-state abbreviation, reversed component order, comma insertion,
and lowercasing. Unlike
unrestricted textual augmentation, these operations have an explicit
invariance contract: they alter formatting or hierarchy expression without
changing denotation. This is important for geography, where a small character
edit can produce another valid place name. Common misspellings and orthographic
variants are therefore admitted only through the alias validation path below,
rather than through unconstrained character corruption.

\subsection{Generate--verify--calibrate alias learning}
\label{sec:aliases}

Invariant views cannot derive knowledge-bearing names such as ``NYC'', ``the Bay
Area'', or ``pink city''. An instruction-tuned model first proposes
abbreviations, nicknames, alternative spellings, and colloquial forms from
$H_e$. An independent verifier evaluates each proposal against the canonical
entity and its hierarchy. We retain
\begin{equation}
 A_e=\{a: a\in G(H_e),\;\operatorname{Accept}(a,e,H_e)=1\},
 \label{eq:alias_filter}
\end{equation}
and remove case-insensitive copies of the canonical name. This generate--verify
separation expands candidate coverage while preventing unverified generations
from entering the supervision set.

Uniform alias weighting improved conditional ranking quality but reduced Top-5
recall: verbose and lower-confidence forms shifted probability mass away from
minimal names. We therefore use calibrated non-uniform weights that prioritize
minimal stylized names, then validated aliases, then other stylized forms. A
smooth prominence-dependent sampler additionally increases the sampling rate for
prominent entities. The resulting
positive distribution can be written as
\begin{equation}
 p(q\mid e)\propto w(s)\,p(t),\quad q=t(s),\quad
 s\in S_e\cup A_e,\;t\in\mathcal{T}.
 \label{eq:view_distribution}
\end{equation}
Alias generation therefore expands the support of the positive distribution,
while sampling calibration bounds its effect on the learned geometry.

\subsection{Ambiguity-aware negative construction}
\label{sec:negatives}

Uniformly sampled negatives are dominated by lexically and geographically
unrelated entities, yielding low-confusability comparisons with limited signal
for geographic disambiguation. We instead construct negatives along two axes.
First, \emph{hierarchy-overlap negatives} share a subset of the query's name and
hierarchy components. For ``San Jose US'', San Jose, Costa Rica preserves the
city name but conflicts with the country constraint, whereas an unrelated
country shares neither component. We stratify candidates by component overlap
and sample round-robin across strata so that high-frequency patterns do not
dominate the constructed-negative distribution.

Second, same-name entities require false-negative control. A less prominent
entity is not automatically wrong; ``Alexandria'' may validly refer to several
places. We use same-name entity $e^-$ as a negative for $e^+$ only when
\begin{equation}
 m_{e^+}+1 \ge \gamma (m_{e^-}+1),
 \label{eq:dominance}
\end{equation}
where $\gamma>1$ is a fixed dominance threshold; we also cap negative exposure
relative to positive frequency. The location-popularity token supplies an
explicit tie-breaker among same-name geographic entities. Together, the
dominance threshold and cap bound the contrastive penalty assigned to valid
secondary interpretations. This design addresses a failure mode of the earlier
objective, which improved the dominant entity's rank by suppressing legitimate
secondary entities. Each positive receives a bounded set of constructed
negatives; other batch positives provide global in-batch negatives.

\subsection{Two-timescale alias memory}
\label{sec:alias_updates}

Aliases have two distinct roles. Query-side aliases supervise a semantic mapping
that transfers across entities. Entity-side aliases in
Equation~\ref{eq:entity_doc} explicitly condition the representation on
entity-specific or newly discovered knowledge. After training with the
\texttt{Also Known As} field, an update $\Delta A_e$ changes only
\begin{equation}
 \mathbf{v}'_e=\operatorname{norm}\!\left(P_d
 h_\theta(p_e,d(e,A_e\cup\Delta A_e))\right),\qquad \theta'=\theta.
 \label{eq:local_update}
\end{equation}
The system re-embeds the affected entity and replaces its index vector; it does
not regenerate training data or update model parameters. This factorization
yields two update timescales: model parameters encode cross-entity regularities,
while entity documents store mutable, entity-specific facts.
Section~\ref{sec:update_benchmark} evaluates the resulting update along three
axes: new-alias acquisition, canonical-form retention, and locality on
unaffected queries. As a concrete intervention,
adding \texttt{SLC} to the Salt Lake City document makes the previously missed
query retrieve the intended entity after re-embedding. We restrict the document
field to validated aliases; including lower-confidence taxonomy abbreviations
reduced retrieval quality in an earlier version.

\subsection{Training-instance assembly}
\label{sec:instance_assembly}

The preceding components define a single training-instance generator. We first
sample an entity using the smoothed prominence-aware distribution, then select a
minimal name, validated alias, or other stylized form using the calibrated
ordering above. An identity-preserving transformation produces the query view.
The positive is the document of the same ontology entity, including its validated
entity-side aliases. We then attach a bounded set of constructed negatives by cycling through
hierarchy-overlap strata and dominance-qualified same-name candidates; positive
documents belonging to other batch items supply global in-batch negatives.
Consequently, lexical proximity alone never makes an entity negative: ontology
identity determines the positive, hierarchy creates targeted contrast, and
Equation~\ref{eq:dominance} gates ambiguous same-name contrast.

This assembly also separates two uses of synthetic text. Generated aliases can
enter query-side supervision only after referent verification, where they teach
a mapping intended to transfer across entities. The entity-side
\texttt{Also Known As} field stores the validated subset as editable content.
Sampling calibration controls how often aliases affect parameter updates,
whereas editing the document changes only one entity vector. The distinction is
important operationally: increasing alias sampling mass can change the global
embedding geometry, while an entity edit is local to the refreshed vector. It
also motivates reporting retrieval quality and editability as related but
different properties of the system.

\subsection{Objective and compact-space training}
\label{sec:model}

We fine-tune an open-source 0.6B-parameter embedding model as a shared-weight,
prompt-asymmetric bi-encoder. For positive pair $(q_i,e_i^+)$ and the union
$\mathcal{N}_i$ of constructed and in-batch negatives, the cached
multiple-negatives objective is
\begin{equation}
 \mathcal{L}_i=-\log\frac{\exp(s(q_i,e_i^+)/\tau)}
 {\exp(s(q_i,e_i^+)/\tau)+\sum_{e\in\mathcal{N}_i}\exp(s(q_i,e)/\tau)}.
\end{equation}
For all reported experiments, we use full fine-tuning for two epochs
(batch size 256, learning rate $10^{-4}$, maximum sequence length 512, and
bfloat16 precision). Frozen 0.6B, 4B, and 8B baselines come from one public
embedding family; adaptation starts from its 0.6B checkpoint, with architecture
and tokenizer unchanged. A Matryoshka objective
\cite{kusupati2022matryoshka} applies the retrieval loss to the leading 64
coordinates $P_d$ in Equation~\ref{eq:local_update}, rather than evaluating a
compact post-hoc projection, thereby aligning training and evaluation. Query and
entity templates are fixed. These are evaluated study settings, not
deployed-system specifications; the exact checkpoint identity is omitted under
organizational disclosure constraints. Controlled comparisons fix
initialization, training budget, the 64-dimensional objective, templates, and
evaluation sets, varying only the stated supervision or document component.

Section~\ref{sec:task_adaptation} varies frozen encoder size; the GeoNames
transfer holds the selected 0.6B checkpoint fixed.

We do not apply unrestricted character-level typo corruption. Alternative
spellings enter supervision only through the same validation path used for
other aliases, preventing arbitrary edits from changing the intended referent.

\section{Experiments}
\label{sec:experiments}

We evaluate four questions: whether the specialized design contributes beyond
standard fine-tuning and encoder scale; how entity-only updates trade alias acquisition against canonical
retention and compare with lexical insertion; whether the selected model
transfers to public GeoNames; and what relevance and engagement evidence is
provided by blinded production comparisons, fixed-query endpoint evaluation,
and a live A/B test.
\subsection{Evaluation protocol}
\label{sec:offline_protocol}

\textbf{Leakage control and split integrity.} The benchmark holds out surface
forms, not entities, from a fixed inventory, and evaluation queries are frozen
before training. After Unicode normalization, case folding, and punctuation and
whitespace normalization, we remove every evaluation-form match from query
supervision, generated aliases, and entity documents. We define near duplicates
by Jaccard similarity $\geq 0.8$ over boundary-padded 3--5-character-gram unions
and keep each resulting group within one split. Filtering is repeated after
alias generation. Consequently, no normalized evaluation form is used as a
training query or document alias, and no near-duplicate group crosses splits.

\textbf{Fixed prompted geo judge.} For entity retrieval, a prompted LLM judge
receives only the extracted location text and the candidate's name, type,
country, city, and administrative hierarchy; geohash and retrieval scores are
excluded. It scores each pair $(q,e)$ from 0 to 4 using the rubric in
Table~\ref{tab:rubric}. The rubric separates the dominant referent from
technically valid but unlikely same-name places, and penalizes a nearby place
that does not denote the query. We aggregate the grades with nDCG@10 and
thresholded Success@5 and precision at the reported cutoffs; unless a stricter
threshold is shown explicitly, grades 3--4 count as relevant. The same metric
implementation and ideal-gain convention in Section~\ref{sec:problem} are used
for every compared variant. Specifically, for each query we pool and
deduplicate the Top-10 entities from all systems in the reported comparison,
judge each pooled query--entity pair, and derive a shared IDCG@10 from that
pool. The judge model, prompt,
decoding configuration, rubric, and input fields are fixed across all compared
retrieval variants. We do not assume its validity from prior work; the blinded
five-annotator study in Section~\ref{sec:human_validation} directly measures its
agreement with human consensus on our geo-specific outputs.

\begin{table}[!ht]
\caption{Geo-relevance rubric, illustrated for the query ``la''.}
\label{tab:rubric}
\small
\begin{tabular}{@{}cll@{}}
\toprule
Score & Interpretation & Example \\
\midrule
4 & Dominant referent & Los Angeles, California \\
3 & Valid, non-dominant name match & La, Ghana \\
2 & Geographically related, not denoted & Long Beach, California \\
1 & Weak or incidental relation & New York, United States \\
0 & No relation & Paris, France \\
\bottomrule
\end{tabular}
\end{table}

\subsection{Task adaptation versus encoder scale}
\label{sec:task_adaptation}

We first test whether pretrained model scale can replace task adaptation. All
variants are evaluated on the same large, fixed production-derived development
set against the same entity inventory. The
prompted LLM judge, its 0--4 rubric,
and the retrieval pipeline are fixed across variants. The frozen checkpoints come from a single open-source embedding family at
0.6B, 4B, and 8B parameters. We apply the same prompts and retrieval path to
each checkpoint without task-specific parameter updates.
Table~\ref{tab:scale_adaptation} compares them with the selected task-adapted
0.6B model. Success@5 ($\geq t$) is the fraction of queries for which at
least one Top-5 entity receives judge grade $t$ or higher; nDCG@10 retains
the graded relevance signal.

\begin{table}[!ht]
\caption{Frozen open-source models versus task adaptation. Success@5 counts
queries with at least one Top-5 result at or above the indicated grade
threshold; bold marks the best result.}
\label{tab:scale_adaptation}
\small
\centering
\setlength{\tabcolsep}{4.8pt}
\begin{tabular}{@{}lrrr@{}}
\toprule
& \multicolumn{2}{c}{Success@5} & \\
\cmidrule(lr){2-3}
Model & $\geq 3$ & $\geq 4$ & nDCG@10 \\
\midrule
Open-source 0.6B & 0.1240 & 0.0956 & 0.4581 \\
Open-source 4B & 0.2075 & 0.1548 & 0.6993 \\
Open-source 8B & 0.2816 & 0.2377 & 0.7124 \\
\textbf{Adapted 0.6B (ours)} & \textbf{0.8931} & \textbf{0.8499} & \textbf{0.9201} \\
\bottomrule
\end{tabular}
\end{table}

Task adaptation raises nDCG@10 by 0.4620 (100.9\%) over the same-size frozen
encoder and by 0.2077 (29.2\%) over the frozen 8B encoder. The Success@5 gaps
are similarly large. Thus, increasing pretrained model scale alone does not
recover the task-specific supervision. Because this benchmark is used for model selection, these are controlled
development-set point estimates rather than held-out test results; GeoNames and
the production evaluations provide external evidence.

\noindent\textbf{Controlled ablation study.}
All unlisted factors are held fixed: initialization, random seed, optimization
schedule, training budget, entity inventory, and evaluation set. The first row is
standard surface-form fine-tuning with the shared encoder and objective; each
subsequent comparison changes only the named intervention.

\begin{table}[t]
\caption{Controlled ablations of supervision and entity representation. In the
first block, each $\Delta$ is relative to the preceding row. In the second, JSON
and validated-only compare with the NL reference; no-geohash compares with
validated-only. Bold denotes the offline best; the dagger marks the deployed variant.}
\label{tab:sequential_ablation}
\small
\centering
\setlength{\tabcolsep}{3.5pt}
\begin{tabularx}{\columnwidth}{@{}Xrr@{}}
\toprule
Configuration & $\Delta$ & nDCG@10 \\
\midrule
\multicolumn{3}{@{}l}{\textit{Cumulative supervision and ambiguity handling}} \\
Standard surface-form fine-tuning & --- & 0.8099 \\
$+$ Initial ambiguity-aware sampling & $+0.0511$ & 0.8610 \\
$+$ Refined ambiguity-aware sampling & $+0.0087$ & 0.8697 \\
$+$ Validated query-side aliases & $+0.0127$ & 0.8824 \\
$+$ Calibrated source sampling & $+0.0067$ & 0.8891 \\
$+$ Popularity-bounded same-name negatives & $+0.0197$ & 0.9088 \\
\addlinespace[2pt]
\multicolumn{3}{@{}l}{\textit{Controlled entity-representation changes}} \\
NL documents with editable aliases (reference) & --- & 0.9133 \\
JSON serialization only & $-0.0092$ & 0.9041 \\
\textbf{Validated-only entity aliases} & $\mathbf{+0.0118}$ & \textbf{0.9251} \\
Validated aliases, no geohashes$\dagger$ & $-0.0050$ & 0.9201 \\
\bottomrule
\end{tabularx}
\end{table}

The cumulative supervision sequence raises nDCG@10 from 0.8099 to 0.9088.
Initial ambiguity-aware sampling gives the largest single increase ($+0.0511$);
validated query aliases add 0.0127, and popularity-bounded same-name negatives
add 0.0197, consistent with their intended roles. Relative to natural-language
documents, JSON serialization reduces nDCG@10 by 0.0092 and validated-only
entity aliases improve it by 0.0118. Removing geohashes from that offline best
costs only 0.0050 (0.54\%) and caused no meaningful endpoint change while
simplifying document updates, so we deployed the no-geohash variant. The first
row is the direct control for ordinary task fine-tuning: the specialized
supervision and selected representation raise nDCG@10 by a further 0.1102, from
0.8099 to 0.9201, under the fixed conditions above. These deltas are conditional
component effects, not training-seed variance estimates.

\subsection{Editable alias memory}
\label{sec:update_benchmark}

We evaluate tens of thousands of held-out alias--entity mappings across tens
of thousands of entities. Their query forms are absent from task training and the
corresponding pre-update entity documents; 3.4\% of unique surfaces map to
multiple geographic entity IDs. Acquisition uses every selected mapping, retention uses
canonical queries available for the updated entities, and locality uses
disjoint alias and canonical controls whose targets are not updated.

For the entity-document update, we add aliases only to the target entity's
\texttt{Also Known As} field and recompute that vector with the encoder and all
other vectors frozen. The lexical update inserts the same mappings into an exact
alias table. Each mechanism is paired with its own pre-update state. We report
target Recall@1 and Recall@5 for acquisition, canonical target Recall@5 change
for retention, and control nDCG@10 change for locality. Confidence intervals use
10,000 entity-level paired bootstrap replicates.

\begin{table}[t]
\caption{Incremental alias acquisition. Each mechanism is compared with its own pre-update state; $\Delta$ cells report entity-level paired-bootstrap 95\% confidence intervals. Unshown disjoint controls are unchanged at four decimals.}
\label{tab:alias_update}
\small
\centering
\setlength{\tabcolsep}{3.5pt}
\begin{tabularx}{\columnwidth}{@{}Xcc@{}}
\toprule
Measure & Before $\to$ after & \shortstack{$\Delta$\\(95\% CI)} \\
\midrule
\multicolumn{3}{@{}l}{\textit{Entity document}} \\
Alias Target R@1 & $0.182 \to 0.644$ & \shortstack{$+0.4616$\\$[+0.4574,+0.4658]$} \\
Alias Target R@5 & $0.287 \to 0.766$ & \shortstack{$+0.4795$\\$[+0.4753,+0.4837]$} \\
Canonical R@5 & $0.981 \to 0.960$ & \shortstack{$-0.0211$\\$[-0.0231,-0.0193]$} \\
\addlinespace[2pt]
\multicolumn{3}{@{}l}{\textit{Lexical alias table}} \\
Alias Target R@1 & $0.022 \to 0.937$ & \shortstack{$+0.9149$\\$[+0.9124,+0.9173]$} \\
Alias Target R@5 & $0.028 \to 0.995$ & \shortstack{$+0.9670$\\$[+0.9655,+0.9685]$} \\
Canonical R@5 & $0.984 \to 0.982$ & \shortstack{$-0.0018$\\$[-0.0023,-0.0014]$} \\
\bottomrule
\end{tabularx}
\end{table}

The entity-document update raises target Recall@1 from 0.1819 to 0.6435 and
target Recall@5 from 0.2869 to 0.7664. Controls remain flat, but canonical
Recall@5 decreases by 0.0211 (95\% CI [$-0.0231$,$-0.0193$]), making the update
local rather than retention-neutral. Exact lexical insertion reaches 0.9369
Recall@1 and 0.9949 Recall@5; its canonical Recall@5 decreases by only 0.0018
(95\% CI [$-0.0023$,$-0.0014$]) and controls remain unchanged. The mechanisms
are therefore complementary: lexical tables suit validated deterministic
mappings, whereas entity documents offer a localized neural update without
encoder retraining, at a larger retention cost.

\subsection{Transfer to a public ontology}
\label{sec:public_ontology}

We test transfer on a frozen GeoNames snapshot released under CC BY 4.0
\cite{geonames2026}. The candidate inventory retains country,
administrative-area, and populated-place entries (feature classes \texttt{A}
and \texttt{P}). Without production fields, each document contains canonical
and ASCII names, feature code, country and administrative hierarchy, and a
population bucket; coordinates remain excluded.

\textbf{Evaluation.} A 3,002,271-query canonical-name slice serves only as a
harness sanity check. Alias evidence uses two fixed sets: known-target metrics
use 860,582 source-linked pairs, whereas pooled graded metrics use 716,322
held-out queries. Every method shares the queries and candidate inventory within
each metric group. Query forms are excluded from the corresponding training
pairs and entity documents. In the source-linked set, 29.5\% of pairs have a
surface associated with multiple entities and are scored against their own
source entity.

Lexical systems share Unicode-aware query/document normalization: Latin
diacritics are removed after decomposition, text is lowercased, and runs of
non-alphanumeric, non-mark characters collapse to spaces. Character retrieval
unions boundary-padded whole-string 3--5-grams, scores them with sublinear
TF--IDF cosine, and drops grams occurring in more than 1\% of documents;
BM25 uses $k_1=1.2$ and $b=0.75$. All systems use the same public name fields
and inventory, produce Top-10 before labels, and exclude each held-out form
from its source entity document.

The proprietary production standardizer cannot be ported to GeoNames without
replacing its inventory and taxonomy-dependent logic. We instead compare exact
and prefix matching, BM25, character $n$-grams, a frozen open-source 0.6B
encoder, and our checkpoint, which is selected on the production-derived
development set and not tuned on GeoNames. Target Recall@1, Target Recall@5, and
MRR@10 follow directly from the source entity identifier and require no judge; MRR@10 is
zero when the target is absent from the Top-10. Pooled judged nDCG@10 and P@1
allow another entity sharing the surface to receive graded relevance.

\begin{table*}[t]
\centering
\caption{GeoNames alias transfer. Known-target metrics use each alias's source entity; pooled metrics allow graded relevance to other valid entities. Bold marks the best result.}
\label{tab:public_ontology}
\small
\setlength{\tabcolsep}{10pt}
\begin{tabular}{@{}lccccc@{}}
\toprule
& \multicolumn{3}{c}{Known source entity} & \multicolumn{2}{c}{Pooled graded relevance} \\
\cmidrule(lr){2-4}\cmidrule(l){5-6}
Method & Target R@1 & Target R@5 & MRR@10 & nDCG@10 & P@1 \\
\midrule
Exact token & 0.0155 & 0.0324 & 0.0227 & 0.1222 & 0.2570 \\
Prefix token & 0.0481 & 0.0805 & 0.0621 & 0.1111 & 0.2251 \\
BM25 & 0.0622 & 0.1246 & 0.0891 & 0.1681 & 0.3137 \\
Character $n$-gram & 0.1064 & \textbf{0.2133} & 0.1525 & 0.2110 & 0.3550 \\
\addlinespace[2pt]
Open-source 0.6B (frozen) & 0.0424 & 0.0851 & 0.0607 & 0.1280 & 0.1356 \\
\textbf{Task-adapted (ours)} & \textbf{0.1165} & 0.2078 & \textbf{0.1557} & \textbf{0.2113} & \textbf{0.3612} \\
\bottomrule
\end{tabular}
\end{table*}

\begin{table*}[t]
\centering
\caption{Character-overlap analysis on GeoNames. Bold marks the higher Target Recall@1 per slice; ours leads for all $J\leq0.50$ bands (79.1\% of pairs) and overall, while character $n$-gram leads for $J>0.50$. Deltas are ours minus character $n$-gram; 95\% CIs use 10,000 query-clustered bootstrap replicates.}
\label{tab:overlap_analysis}
\small
\setlength{\tabcolsep}{5.0pt}
\begin{tabular}{@{}lrrrrr@{}}
\toprule
Slice & \shortstack{Pairs\\(share)} & Char R@1 & Ours R@1
& \shortstack{$\Delta$R@1\\(95\% CI)}
& \shortstack{$\Delta$R@5\\(95\% CI)} \\
\midrule
$J=0$ & \shortstack[r]{236,608\\(27.5\%)} & 0.0000 & \textbf{0.0163} & \shortstack[r]{$+0.0163$\\$[.0158,.0169]$} & \shortstack[r]{$+0.0408$\\$[.0399,.0417]$} \\
$0<J\leq0.25$ & \shortstack[r]{202,469\\(23.5\%)} & 0.0101 & \textbf{0.0469} & \shortstack[r]{$+0.0368$\\$[.0358,.0378]$} & \shortstack[r]{$+0.0784$\\$[.0769,.0799]$} \\
$0.25<J\leq0.50$ & \shortstack[r]{241,499\\(28.1\%)} & 0.1177 & \textbf{0.1630} & \shortstack[r]{$+0.0453$\\$[.0437,.0470]$} & \shortstack[r]{$-0.0055$\\$[-.0077,-.0033]$} \\
$J>0.50$ & \shortstack[r]{180,006\\(20.9\%)} & \textbf{0.3394} & 0.2642 & \shortstack[r]{$-0.0752$\\$[-.0774,-.0730]$} & \shortstack[r]{$-0.1607$\\$[-.1632,-.1582]$} \\
Overall & \shortstack[r]{860,582\\(100\%)} & 0.1064 & \textbf{0.1165} & \shortstack[r]{$+0.0101$\\$[.0094,.0108]$} & \shortstack[r]{$-0.0055$\\$[-.0064,-.0046]$} \\
\bottomrule
\end{tabular}
\end{table*}

For known-target metrics, the aggregate task-adapted-minus-character-$n$-gram
differences are $+0.0101$ at Recall@1 (95\% CI
[$+0.0094$,$+0.0108$]), $-0.0055$ at Recall@5
([$-0.0064$,$-0.0046$]), and $+0.0032$ at MRR@10
([$+0.0025$,$+0.0039$]). The intervals use 10,000 query-clustered paired
bootstrap replicates and none spans zero.

Fixed overlap bands localize this trade-off. Adaptation significantly improves
Target Recall@1 in all $J\leq0.50$ bands (79.1\% of pairs), with the gain
rising from 0.0163 at $J=0$ to 0.0453 for moderate overlap; character
$n$-grams win for $J>0.50$. At Target Recall@5, gains in the zero- and
low-overlap bands are outweighed by the high-overlap loss, producing the
aggregate $-0.0055$. Both systems improve in absolute Recall@1 as overlap
rises, so the result indicates lower, not zero, dependence on surface overlap.
GeoNames is a transfer benchmark, not a production-selection proxy: the
overlap analysis diagnoses lexical regimes, while deployment selection rests on
the production-derived, human, endpoint, and live evidence below.

On pooled judged point estimates, adaptation improves over the frozen encoder
by 65.1\% in nDCG@10 and 166.4\% in P@1, exceeds exact, prefix, and BM25 on
both measures, and narrowly leads character $n$-grams by 0.0003 nDCG@10 and
0.0062 P@1. On the canonical sanity slice, exact matching and the adapted model
reach 0.9699 and 0.9885 P@1.

\subsection{Blinded human evaluation and judge validation}
\label{sec:human_validation}

The 50-query challenge set comprises 30 high-frequency production location
queries, 10 queries selected by product managers from quality tickets filed
before the present evaluation, and 10 queries sampled at random from historical
search impressions. We intentionally restrict the set to queries for which the
systems' Top-1 entities differ. It targets actionable cases rather than
estimating a traffic-wide effect. Five internal domain experts from engineering and product management,
all familiar with people search and geographic standardization, independently
graded both outputs for every query on the 0--4 scale in Table~\ref{tab:rubric}. The 100 query--document pairs were
presented in random order without system identifiers, yielding 500 ratings.

For each output, grade@1 is the mean of its five ratings. We call an output
relevant when at least three annotators assign grade 3 or 4. Mean-grade
differences use the 50 complete production--embedding pairs; the confidence
interval is obtained by paired query-level bootstrap and the $p$-value by a
paired $t$-test. Relevant P@1 uses an exact McNemar test. Win/tie/loss compares
the two mean grades within each query and uses an exact sign test after removing
ties.

For judge validation, the human consensus for item $i$ is the median of its
five ordinal ratings,
$h_i^{\mathrm{consensus}}=\operatorname{median}(h_{i1},\ldots,h_{i5})$.
Spearman's $\rho$ measures rank association between the judge grade and this
consensus; weighted Cohen's $\kappa$ measures agreement while accounting for
the distance between ordinal grades.

\begin{table}[t]
\caption{Blinded Top-1 comparison on the stratified production challenge set.
W/T/L denotes embedding wins, ties, and losses.}
\label{tab:human_results}
\small
\renewcommand{\arraystretch}{1.08}
\begin{tabularx}{\columnwidth}{@{}Xrrr@{}}
\toprule
Metric & Prod. & Emb. & Comparison \\
\midrule
Average grade@1 (0--4) & 1.948 & 2.244 & $+0.296$ \\
Relevant P@1 & \shortstack[r]{28.0\%\\(14/50)} & \shortstack[r]{46.0\%\\(23/50)} & $+18.0$ pp \\
Embedding W/T/L & --- & --- & 26 / 14 / 10 \\
\bottomrule
\end{tabularx}
\parbox{\columnwidth}{\footnotesize\textit{Paired inference.} Grade@1:
95\% CI [$+0.136$, $+0.456$], $p=0.001$; relevant P@1: [$+6.0$, $+30.0$] pp,
$p=0.012$; W/T/L: exact sign test $p=0.011$.}
\end{table}

The embedding system raises average grade@1 by 0.296 ($p=0.001$) and relevant P@1 by 18.0 percentage
points, from 28.0\% to 46.0\%. Ten queries are relevant only for embedding and
one only for production ($p=0.012$, exact McNemar); graded outcomes give
26/14/10 wins/ties/losses ($p=0.011$). Thus both graded and thresholded human
measures favor embedding on this stratified Top-1 challenge set.

\paragraph{People-result human audit.}
Three experts independently scored blinded Top-1 outputs for 50 paired endpoint
queries (100 items; 300 ratings). Median grade defines consensus, grades 3--4
define relevance, and wins compare paired consensus grades.

\begin{table}[t]
\caption{Blinded endpoint audit and judge validation against median human
consensus.}
\label{tab:human_audits}
\small
\centering
\textit{(a) Endpoint Top-1 comparison}\\[2pt]
\begin{tabularx}{\columnwidth}{@{}Xrrr@{}}
\toprule
Metric & Control & Treat. & $\Delta$/overall \\
\midrule
Mean consensus grade@1 & 3.00 & 3.04 & $+0.04$ \\
Relevant P@1 ($\geq3$) & 72\% & 74\% & $+2$ pp \\
Treatment/Control/Ties & --- & --- & 14 / 12 / 24 \\
\bottomrule
\end{tabularx}

\vspace{3pt}
\textit{(b) Agreement by evaluation distribution}\\[2pt]
\begin{tabularx}{\columnwidth}{@{}Xrrrrr@{}}
\toprule
Audit & Items & Exp. & $\alpha$ & QWK & $\rho$ \\
\midrule
Production geo & 100 & 5 & 0.712 & 0.761 & 0.781 \\
People results & 100 & 3 & 0.790 & 0.680 & 0.730 \\
\bottomrule
\end{tabularx}
\parbox{\columnwidth}{\footnotesize Panel (a): excluding 24 ties, the
two-sided exact sign test gives $p=0.845$. Exp. denotes experts; QWK and
Spearman's $\rho$ compare each judge with median human consensus.}
\end{table}

Treatment and control have mean consensus grades of 3.04 and 3.00 and relevant
P@1 of 74\% and 72\%. Treatment wins 14 queries, control wins 12, and 24 tie
($p=0.845$); this supports similar sampled Top-1 relevance, not equivalence.
For production geo, expert agreement is $\alpha=0.712$ and judge alignment is
QWK $=0.761$, $\rho=0.781$; for people results, the corresponding values are
0.790, 0.680, and 0.730. Exact and within-one-level judge agreement on people
results are 64\% and 93\%; binary agreement at grade 3 is 92\%
($\kappa=0.83$).

\subsection{Online relevance and engagement evaluation}
\label{sec:online}

Paired production-endpoint calls use traffic-derived frequent queries and
non-canonical-location queries with randomly sampled member identifiers.
Control uses the taxonomy standardizer; treatment enables Top-1 or Top-5
geo-embedding filters with the remaining stack fixed. The human-validated
people-result judge in Table~\ref{tab:human_audits} grades the returned Top-10
people results. Geo Top-$k$ denotes filter depth, not people-result rank. These
fixed-query calls measure serving relevance rather than live-user effects.

\begin{table}[t]
\caption{Fixed-query endpoint relevance. Values are mean judged grade@10;
Geo Top-$k$ is geographic retrieval depth. Relative changes are computed from
unrounded means. Descriptive point estimates only; no confidence intervals are
available.}
\label{tab:endpoint_replay}
\small
\renewcommand{\arraystretch}{1.12}
\begin{tabularx}{\columnwidth}{@{}Xrrr@{}}
\toprule
Query set & Control & Geo Top-1 & Geo Top-5 \\
\midrule
Traffic-derived frequent queries & 2.7289 & \shortstack[r]{2.7390\\($+0.4\%$)} & \shortstack[r]{2.7223\\($-0.2\%$)} \\
Non-canonical locations & 3.3311 & \shortstack[r]{\textbf{3.3997}\\\textbf{($+2.1\%$)}} & \shortstack[r]{\textbf{3.4959}\\\textbf{($+5.0\%$)}} \\
\bottomrule
\end{tabularx}
\end{table}

Frequent-query estimates stay within 0.4\% of control; non-canonical estimates
increase by 2.1\% at Top-1 and 5.0\% at Top-5. Without query-level intervals,
we make no significance or non-inferiority claim. Separately, a one-week, 50/50
member-randomized A/B test detects no regression in top-level engagement and
search-health guardrails, including query volume and search success rate; exact
values are withheld.

\noindent\textbf{Limitations.}
GeoNames differs from production, and its aliases may occur in base-model
pretraining. Its pooled judged results are point estimates, while paired
intervals cover only known-target comparison with character $n$-grams. The
update benchmark tests local edits rather than globally novel knowledge.
Finite human audits do not establish judge validity across all regions,
languages, query types, or ranks; character overlap diagnoses lexical evidence,
not semantic reasoning. Fixed-query endpoint results lack randomization and
intervals, while the live test establishes only an engagement guardrail.

\section{Discussion}
\label{sec:deployment}

Ordinary task fine-tuning reaches 0.8099 nDCG@10 versus 0.9201 for the selected
task-adapted system under fixed conditions; Table~\ref{tab:sequential_ablation} thus
reports conditional effects under one seed. On GeoNames, adaptation improves
Target Recall@1 for $J\leq0.50$ (79.1\% of pairs), whereas character $n$-grams
dominate at high overlap, making aggregate Recall@5 a regime-composition
result. This public baseline does not change the production role: the learned
standardizer replaces the incumbent taxonomy-based path.

Other evaluations also answer distinct questions. Lexical insertion is strongest
for deterministic mappings; entity documents enable neural updates but reduce
canonical retention. The challenge set measures Top-1 disagreements, endpoint
gains are descriptive, and the live experiment supplies only an engagement
guardrail. These estimands should not be conflated.

\section{Conclusion}
\label{sec:conclusion}

We formulate geographic standardization as graded, set-valued retrieval.
Fixed-seed development ablations show gains beyond ordinary task fine-tuning
and encoder scale, tracing them to ambiguity-aware sampling, verified aliases,
and mutable entity documents.

On GeoNames, adaptation leads Target Recall@1 and MRR@10 overall and Recall@1
for $J\leq0.50$; character $n$-gram leads on high-overlap forms and aggregate
Target Recall@5. The gain is therefore regime-dependent.

Human relevant P@1 rises from 28.0\% to 46.0\%; endpoint estimates improve on
non-canonical queries with no live engagement regression. Together, these
results support replacing taxonomy-based standardization with task-adapted
retrieval.

\section{Ethical and Responsible AI Considerations}
\label{sec:ethics}

The geographic model is non-personalized: it resolves location phrases against a
curated ontology and does not encode, profile, or rank people, or use individual
sensitive attributes as model features. The member-count signal is aggregated at
the location level and used only to order geographic candidates for same-name
disambiguation; it is never used to rank people. Reported production results are
aggregate changes rather than individual query or member records.

The audit of the stratified production challenge set used five internal domain
experts from
engineering and product management; the people-result audit used three internal
expert raters. All raters were familiar with people search and used the
predefined ordinal relevance rubric. Outputs were randomized and system-blinded,
and every expert independently rated every item in the corresponding audit.
The expert audits and member-randomized experiment followed the organization's
applicable experimentation, privacy, and human-participant review requirements.

Safeguards are applied at the geographic-entity level. Popularity is bucketed,
and dominance gating with a per-entity cap constrains its use in same-name
negative construction. Aliases are validated before entering training or entity
documents. Public-catalog transfer, the audit of the stratified production challenge set,
and the people-result audit are reported separately. The challenge set's purposive
sampling and limited scope are disclosed rather than used as evidence of
production-wide, regional, or linguistic quality.

Our evaluation draws from a global location inventory and is not restricted to
a particular language or region. Aggregate metrics nevertheless do not establish
uniform performance across region--language strata. Future evaluation should
report region-by-language slices, false-broadening rates, and performance for
low-population entities, with human review of a stratified sample, before making
parity claims.

\balance
\bibliographystyle{ACM-Reference-Format}
\bibliography{refs}

@inproceedings{huang2013dssm,
  title={Learning Deep Structured Semantic Models for Web Search Using Clickthrough Data},
  author={Huang, Po-Sen and He, Xiaodong and Gao, Jianfeng and Deng, Li and Acero, Alex and Heck, Larry},
  booktitle={Proceedings of the 22nd ACM International Conference on Information and Knowledge Management},
  pages={2333--2338},
  year={2013}
}

@inproceedings{karpukhin2020dpr,
  title={Dense Passage Retrieval for Open-Domain Question Answering},
  author={Karpukhin, Vladimir and O{\u{g}}uz, Barlas and Min, Sewon and Lewis, Patrick and Wu, Ledell and Edunov, Sergey and Chen, Danqi and Yih, Wen-tau},
  booktitle={Proceedings of the 2020 Conference on Empirical Methods in Natural Language Processing},
  pages={6769--6781},
  year={2020}
}

@inproceedings{reimers2019sbert,
  title={Sentence-{BERT}: Sentence Embeddings Using Siamese {BERT}-Networks},
  author={Reimers, Nils and Gurevych, Iryna},
  booktitle={Proceedings of the 2019 Conference on Empirical Methods in Natural Language Processing and the 9th International Joint Conference on Natural Language Processing},
  pages={3982--3992},
  year={2019}
}

@inproceedings{wu2020blink,
  title={Scalable Zero-shot Entity Linking with Dense Entity Retrieval},
  author={Wu, Ledell and Petroni, Fabio and Josifoski, Martin and Riedel, Sebastian and Zettlemoyer, Luke},
  booktitle={Proceedings of the 2020 Conference on Empirical Methods in Natural Language Processing},
  pages={6397--6407},
  year={2020}
}

@inproceedings{weissenbacher2019semeval,
  title={{SemEval}-2019 Task 12: Toponym Resolution in Scientific Papers},
  author={Weissenbacher, Davy and Magge, Arjun and O'Connor, Karen and Scotch, Matthew and Gonzalez-Hernandez, Graciela},
  booktitle={Proceedings of the 13th International Workshop on Semantic Evaluation},
  pages={907--916},
  year={2019}
}

@inproceedings{gritta2018toponym,
  title={Which {Melbourne}? Augmenting Geocoding with Maps},
  author={Gritta, Milan and Pilehvar, Mohammad Taher and Collier, Nigel},
  booktitle={Proceedings of the 56th Annual Meeting of the Association for Computational Linguistics},
  year={2018}
}

@inproceedings{zhang2023geonorm,
  title={Improving Toponym Resolution with Better Candidate Generation, Transformer-based Reranking, and Two-Stage Resolution},
  author={Zhang, Zeyu and Bethard, Steven},
  booktitle={Proceedings of the 12th Joint Conference on Lexical and Computational Semantics},
  pages={48--60},
  publisher={Association for Computational Linguistics},
  doi={10.18653/v1/2023.starsem-1.6},
  year={2023}
}

@inproceedings{li2023geolm,
  title={{GeoLM}: Empowering Language Models for Geospatially Grounded Language Understanding},
  author={Li, Zekun and Zhou, Wenxuan and Chiang, Yao-Yi and Chen, Muhao},
  booktitle={Proceedings of the 2023 Conference on Empirical Methods in Natural Language Processing},
  pages={5227--5240},
  year={2023}
}

@inproceedings{masis2024earth,
  title={Where on Earth Do Users Say They Are?: Geo-Entity Linking for Noisy Multilingual User Input},
  author={Masis, Tessa and O'Connor, Brendan},
  booktitle={Proceedings of the Sixth Workshop on Natural Language Processing and Computational Social Science},
  pages={86--98},
  publisher={Association for Computational Linguistics},
  doi={10.18653/v1/2024.nlpcss-1.7},
  year={2024}
}

@inproceedings{xiong2021ance,
  title={Approximate Nearest Neighbor Negative Contrastive Learning for Dense Text Retrieval},
  author={Xiong, Lee and Xiong, Chenyan and Li, Ye and Tang, Kwok-Fung and Liu, Jialin and Bennett, Paul and Ahmed, Junaid and Overwijk, Arnold},
  booktitle={International Conference on Learning Representations},
  year={2021}
}

@inproceedings{kusupati2022matryoshka,
  title={Matryoshka Representation Learning},
  author={Kusupati, Aditya and Bhatt, Gantavya and Rege, Aniket and Wallingford, Matthew and Sinha, Aditya and Ramanujan, Vivek and Howard-Snyder, William and Chen, Kaifeng and Kakade, Sham and Jain, Prateek and Farhadi, Ali},
  booktitle={Advances in Neural Information Processing Systems},
  year={2022}
}

@inproceedings{le2026sage,
  title={{SAGE}: Scalable {AI} Governance \& Evaluation},
  author={Le, Benjamin H. and Lu, Xueying and Stern, Nicholas and Liu, Wenqiong and Lapchuk, Igor and Li, Xiang and Zheng, Baofen and Rosenberg, Kevin and Huang, Jiewen and Zhang, Zhe and Cabangbang, Abraham and Wagle, Satej Milind and Shen, Jianqiang and Muthuregunathan, Raghavan and Gupta, Abhinav and Teoh, Mathew and Kirk, Andrew J. N. and Kwan, Thomas and Wu, Jingwei and Zhang, Wenjing},
  booktitle={Proceedings of the 32nd ACM SIGKDD Conference on Knowledge Discovery and Data Mining V.2},
  publisher={ACM},
  doi={10.1145/3770855.3818476},
  year={2026}
}

@inproceedings{faggioli2023perspectives,
  title={Perspectives on Large Language Models for Relevance Judgment},
  author={Faggioli, Guglielmo and Dietz, Laura and Clarke, Charles L. A. and Demartini, Gianluca and Hagen, Matthias and Hauff, Claudia and Kando, Noriko and Kanoulas, Evangelos and Potthast, Martin and Stein, Benno and Wachsmuth, Henning},
  booktitle={Proceedings of the 2023 ACM SIGIR International Conference on Theory of Information Retrieval},
  pages={39--50},
  publisher={ACM},
  doi={10.1145/3578337.3605136},
  year={2023}
}

@inproceedings{thomas2024searcher,
  title={Large Language Models Can Accurately Predict Searcher Preferences},
  author={Thomas, Paul and Spielman, Seth and Craswell, Nick and Mitra, Bhaskar},
  booktitle={Proceedings of the 47th International ACM SIGIR Conference on Research and Development in Information Retrieval},
  pages={1930--1940},
  publisher={ACM},
  doi={10.1145/3626772.3657707},
  year={2024}
}

@article{upadhyay2024umbrela,
  title={{UMBRELA}: {UM}brela Is the (Open-Source Reproduction of the) {B}ing Relevance Assessor},
  author={Upadhyay, Shivani and Pradeep, Ronak and Thakur, Nandan and Craswell, Nick and Lin, Jimmy},
  journal={arXiv preprint arXiv:2406.06519},
  year={2024}
}

@inproceedings{froebe2025assessors,
  title={Large Language Model Relevance Assessors Agree With One Another More Than With Human Assessors},
  author={Fr{\"o}be, Maik and Parry, Andrew and Schlatt, Ferdinand and MacAvaney, Sean and Stein, Benno and Potthast, Martin and Hagen, Matthias},
  booktitle={Proceedings of the 48th International ACM SIGIR Conference on Research and Development in Information Retrieval},
  publisher={ACM},
  doi={10.1145/3726302.3730218},
  year={2025}
}

@inproceedings{takehi2025lara,
  title={{LLM}-Assisted Relevance Assessments: When Should We Ask {LLM}s for Help?},
  author={Takehi, Rikiya and Voorhees, Ellen M. and Sakai, Tetsuya and Soboroff, Ian},
  booktitle={Proceedings of the 48th International ACM SIGIR Conference on Research and Development in Information Retrieval},
  pages={95--105},
  publisher={ACM},
  doi={10.1145/3726302.3729916},
  year={2025}
}

@inproceedings{hashemi2024rubric,
  title={{LLM}-Rubric: A Multidimensional, Calibrated Approach to Automated Evaluation of Natural Language Texts},
  author={Hashemi, Helia and Eisner, Jason and Rosset, Corby and Van Durme, Benjamin and Kedzie, Chris},
  booktitle={Proceedings of the 62nd Annual Meeting of the Association for Computational Linguistics},
  pages={13806--13834},
  publisher={Association for Computational Linguistics},
  doi={10.18653/v1/2024.acl-long.745},
  year={2024}
}

@inproceedings{geyik2018talent,
  title={Talent Search and Recommendation Systems at {LinkedIn}: Practical Challenges and Lessons Learned},
  author={Geyik, Sahin Cem and Guo, Qi and Hu, Bo and Ozcaglar, Cagri and Thakkar, Ketan and Wu, Xianren and Kenthapadi, Krishnaram},
  booktitle={Proceedings of the 41st International ACM SIGIR Conference on Research and Development in Information Retrieval},
  pages={1353--1354},
  publisher={ACM},
  doi={10.1145/3209978.3210205},
  year={2018}
}

@inproceedings{gupta2025people,
  title={Retrieval for Semantic People Search},
  author={Gupta, Rupesh and Zheng, Chujie and Li, Haojun},
  booktitle={Proceedings of the 48th International ACM SIGIR Conference on Research and Development in Information Retrieval},
  pages={4229--4233},
  publisher={ACM},
  doi={10.1145/3726302.3731938},
  year={2025}
}

@article{borisyuk2026semantic,
  title={Semantic Search at {LinkedIn}},
  author={Borisyuk, Fedor and Vasudevan, Sriram and Wu, Muchen and Li, Guoyao and Le, Benjamin and others},
  journal={arXiv preprint arXiv:2602.07309},
  url={https://arxiv.org/abs/2602.07309},
  year={2026}
}

@inproceedings{zhang2024geoplace,
  title={Improving Toponym Resolution by Predicting Attributes to Constrain Geographical Ontology Entries},
  author={Zhang, Zeyu and Laparra, Egoitz and Bethard, Steven},
  booktitle={Proceedings of the 2024 Conference of the North American Chapter of the Association for Computational Linguistics: Human Language Technologies (Short Papers)},
  pages={35--44},
  publisher={Association for Computational Linguistics},
  doi={10.18653/v1/2024.naacl-short.3},
  year={2024}
}

@inproceedings{li2022spabert,
  title={{SpaBERT}: A Pretrained Language Model from Geographic Data for Geo-Entity Representation},
  author={Li, Zekun and Kim, Jina and Chiang, Yao-Yi and Chen, Muhao},
  booktitle={Findings of the Association for Computational Linguistics: EMNLP 2022},
  pages={2757--2769},
  publisher={Association for Computational Linguistics},
  doi={10.18653/v1/2022.findings-emnlp.200},
  year={2022}
}

@inproceedings{nakatani2025poi,
  title={A Text Embedding Model with Contrastive Example Mining for Point-of-Interest Geocoding},
  author={Nakatani, Hibiki and Teranishi, Hiroki and Higashiyama, Shohei and Sawada, Yuya and Ouchi, Hiroki and Watanabe, Taro},
  booktitle={Proceedings of the 31st International Conference on Computational Linguistics},
  pages={7279--7291},
  publisher={Association for Computational Linguistics},
  year={2025}
}

@inproceedings{gillick2019entity,
  title={Learning Dense Representations for Entity Retrieval},
  author={Gillick, Daniel and Kulkarni, Sayali and Lansing, Larry and Presta, Alessandro and Baldridge, Jason and Ie, Eugene and Garcia-Olano, Diego},
  booktitle={Proceedings of the 23rd Conference on Computational Natural Language Learning},
  pages={528--537},
  publisher={Association for Computational Linguistics},
  doi={10.18653/v1/K19-1049},
  year={2019}
}

@inproceedings{sung2020biosyn,
  title={Biomedical Entity Representations with Synonym Marginalization},
  author={Sung, Mujeen and Jeon, Hwisang and Lee, Jinhyuk and Kang, Jaewoo},
  booktitle={Proceedings of the 58th Annual Meeting of the Association for Computational Linguistics},
  pages={3641--3650},
  publisher={Association for Computational Linguistics},
  doi={10.18653/v1/2020.acl-main.335},
  year={2020}
}

@inproceedings{liu2021sapbert,
  title={Self-Alignment Pretraining for Biomedical Entity Representations},
  author={Liu, Fangyu and Shareghi, Ehsan and Meng, Zaiqiao and Basaldella, Marco and Collier, Nigel},
  booktitle={Proceedings of the 2021 Conference of the North American Chapter of the Association for Computational Linguistics: Human Language Technologies},
  pages={4228--4238},
  publisher={Association for Computational Linguistics},
  doi={10.18653/v1/2021.naacl-main.334},
  year={2021}
}

@inproceedings{rucker2025verbalized,
  title={Evaluating Design Decisions for Dual Encoder-based Entity Disambiguation},
  author={R{\"u}cker, Susanna and Akbik, Alan},
  booktitle={Proceedings of the 63rd Annual Meeting of the Association for Computational Linguistics},
  pages={15685--15701},
  publisher={Association for Computational Linguistics},
  doi={10.18653/v1/2025.acl-long.764},
  year={2025}
}

@inproceedings{dai2023promptagator,
  title={Promptagator: Few-shot Dense Retrieval From 8 Examples},
  author={Dai, Zhuyun and Zhao, Vincent Y. and Ma, Ji and Luan, Yi and Ni, Jianmo and Lu, Jing and Bakalov, Anton and Guu, Kelvin and Hall, Keith B. and Chang, Ming-Wei},
  booktitle={International Conference on Learning Representations},
  year={2023}
}

@inproceedings{qu2021rocketqa,
  title={{RocketQA}: An Optimized Training Approach to Dense Passage Retrieval for Open-Domain Question Answering},
  author={Qu, Yingqi and Ding, Yuchen and Liu, Jing and Liu, Kai and Ren, Ruiyang and Zhao, Wayne Xin and Dong, Daxiang and Wu, Hua and Wang, Haifeng},
  booktitle={Proceedings of the 2021 Conference of the North American Chapter of the Association for Computational Linguistics: Human Language Technologies},
  pages={5835--5847},
  publisher={Association for Computational Linguistics},
  doi={10.18653/v1/2021.naacl-main.466},
  year={2021}
}

@inproceedings{logeswaran2019zero,
  title={Zero-Shot Entity Linking by Reading Entity Descriptions},
  author={Logeswaran, Lajanugen and Chang, Ming-Wei and Lee, Kenton and Toutanova, Kristina and Devlin, Jacob and Lee, Honglak},
  booktitle={Proceedings of the 57th Annual Meeting of the Association for Computational Linguistics},
  pages={3449--3460},
  publisher={Association for Computational Linguistics},
  doi={10.18653/v1/P19-1335},
  year={2019}
}

@inproceedings{lewis2020rag,
  title={Retrieval-Augmented Generation for Knowledge-Intensive {NLP} Tasks},
  author={Lewis, Patrick and Perez, Ethan and Piktus, Aleksandra and Petroni, Fabio and Karpukhin, Vladimir and Goyal, Naman and K{\"u}ttler, Heinrich and Lewis, Mike and Yih, Wen-tau and Rockt{\"a}schel, Tim and Riedel, Sebastian and Kiela, Douwe},
  booktitle={Advances in Neural Information Processing Systems},
  volume={33},
  year={2020}
}

@inproceedings{kim2024dynamicer,
  title={{DynamicER}: Resolving Emerging Mentions to Dynamic Entities for {RAG}},
  author={Kim, Jinyoung and Ko, Dayoon and Kim, Gunhee},
  booktitle={Proceedings of the 2024 Conference on Empirical Methods in Natural Language Processing},
  pages={13752--13770},
  publisher={Association for Computational Linguistics},
  doi={10.18653/v1/2024.emnlp-main.762},
  year={2024}
}

@misc{geonames2026,
  author={{GeoNames}},
  title={GeoNames Gazetteer Data Export},
  year={2026},
  howpublished={\url{https://download.geonames.org/export/dump/}},
  note={CC BY 4.0; accessed July 30, 2026}
}

\end{document}